\documentclass[10pt,twocolumn,letterpaper]{article}

\usepackage[dvipsnames,table]{xcolor}
\usepackage{iccv}
\usepackage{times}
\usepackage{epsfig}
\usepackage{graphicx}
\usepackage{amsmath}
\usepackage{amssymb}
\usepackage{enumitem}
\usepackage{threeparttable}
\usepackage{bbm}
\usepackage{pifont}%
\usepackage[export]{adjustbox}
\usepackage{makecell}

\usepackage{appendix}
\usepackage{booktabs}
\usepackage{etoolbox}
\usepackage{booktabs,tabulary,multirow}
\usepackage[caption=false]{subfig}
\usepackage{textcomp}
\usepackage{listings}

\usepackage[T1]{fontenc}
\usepackage[utf8]{inputenc}

\definecolor{background}{RGB}{200, 200, 200}
\definecolor{string}{RGB}{230, 219, 116}
\definecolor{comment}{RGB}{117, 113, 94}
\definecolor{normal}{RGB}{0, 0, 0}
\definecolor{identifier}{RGB}{166, 226, 46}

\usepackage[T1]{fontenc}
\usepackage[utf8]{inputenc}

\definecolor{citecolor}{RGB}{34,139,34}  %
\usepackage[pagebackref=true,breaklinks=true,letterpaper=true,colorlinks,
  citecolor=citecolor,bookmarks=false]{hyperref}
\usepackage{cleveref}
\usepackage{mathtools}

\iccvfinalcopy %

\newcommand{\cmark}{\ding{51}}%
\newcommand{\gxmark}{\textcolor{black!20}{\ding{55}}}%
\newcommand{\gray}[1]{\textcolor{gray}{#1}}%
\newcommand{\tm}{\textminus}%
\newtoggle{comments}
\togglefalse{comments}
\newcommand{\change}[1]{#1}
\iftoggle{comments}{%
    \newcommand{\deva}[1]{{\leavevmode\color{NavyBlue}[Deva: #1]}}
    \newcommand{\achal}[1]{{\leavevmode\color{Emerald}[Achal: #1]}}
    \newcommand{\ross}[1]{{\leavevmode\color{RubineRed}[Ross: #1]}}
    \newcommand{\alex}[1]{{\leavevmode\color{Mulberry}[Alex: #1]}}
    \newcommand{\piotr}[1]{{\leavevmode\color{Peach}[Piotr: #1]}}
}{%
    \newcommand{\deva}[1]{}
    \newcommand{\achal}[1]{}
    \newcommand{\ross}[1]{}
    \newcommand{\alex}[1]{}
    \newcommand{\piotr}[1]{}
}
\newcommand{\smallsec}[1]{\vspace{0.5ex}\noindent{\textbf{#1}}}
\DeclarePairedDelimiter\norm{\lVert}{\rVert}%

\newcommand{\apr}{AP$_\textrm{r}$\xspace}
\newcommand{\apc}{AP$_\textrm{c}$\xspace}
\newcommand{\apf}{AP$_\textrm{f}$\xspace}
\newcommand{\appool}{AP$^{\text{Pool}}$\xspace}
\newcommand{\appoolr}{AP$^{\text{Pool}}_\text{r}$\xspace}
\newcommand{\appoolc}{AP$^{\text{Pool}}_\text{c}$\xspace}
\newcommand{\appoolf}{AP$^{\text{Pool}}_\text{f}$\xspace}
\newcommand{\oldap}{AP$^\text{Old}$\xspace}
\newcommand{\fixedap}{AP$^\text{Fixed}$\xspace}

\definecolor{demphcolor}{RGB}{90,90,90}
\newcommand{\demph}[1]{\textcolor{demphcolor}{#1}}
\newcommand{\dt}[1]{\fontsize{6pt}{0.1em}\selectfont \demph{(#1)}}

\newcommand{\bd}[1]{\textbf{#1}}
\newcommand{\app}{\raise.17ex\hbox{$\scriptstyle\sim$}}

\newcolumntype{x}[1]{>{\centering\arraybackslash}p{#1pt}}
\newlength\savewidth
\newcommand{\tablestyle}[2]{\setlength{\tabcolsep}{#1}\renewcommand{\arraystretch}{#2}\centering\footnotesize}

\newcommand{\tdb}[1]{\scriptsize\rlap{\bd{ \dt{#1}}}}
\newcommand{\td}[1]{\scriptsize\rlap{ \dt{#1}}}

\begin{document}

\title{Evaluating Large-Vocabulary Object Detectors:\\ The Devil is in the Details\vspace{-3mm}}

\author{
Achal Dave$^{1,2}$ \quad Piotr Doll\'ar$^2$ \quad Deva Ramanan$^1$ \quad Alexander Kirillov$^2$ \quad Ross Girshick$^2$ \vspace{.5em} \\
$^1$Carnegie Mellon University  \qquad $^2$Facebook AI Research (FAIR)
}

\maketitle

\begin{abstract}
By design, average precision (AP) for object detection aims to treat all classes independently: AP is computed \emph{independently} per category and averaged.
On one hand, this is desirable as it treats all classes equally.
On the other hand, it ignores cross-category confidence calibration, a key property in real-world use cases.
Unfortunately, under important conditions (\ie, large vocabulary, high instance counts) the default implementation of AP is neither category independent, nor does it directly reward properly calibrated detectors.
In fact, we show that on LVIS the default implementation produces a \emph{gameable} metric, where a simple, unintuitive re-ranking policy can improve AP by a large margin.
To address these limitations, we introduce two complementary metrics.
First, we present a simple fix to the default AP implementation, ensuring that it is independent across categories as originally intended.
We benchmark recent LVIS detection advances and find that many reported gains do not translate to improvements under our new evaluation, suggesting recent improvements may arise from difficult to interpret changes to cross-category rankings.
Given the importance of reliably benchmarking cross-category rankings, we consider a pooled version of AP (\appool) that rewards properly calibrated detectors by directly comparing cross-category rankings. Finally, we revisit classical approaches for calibration and find that explicitly calibrating detectors improves state-of-the-art on \appool{} by 1.7 points.
\end{abstract}

\begin{figure}[h]
    \includegraphics[width=\linewidth]{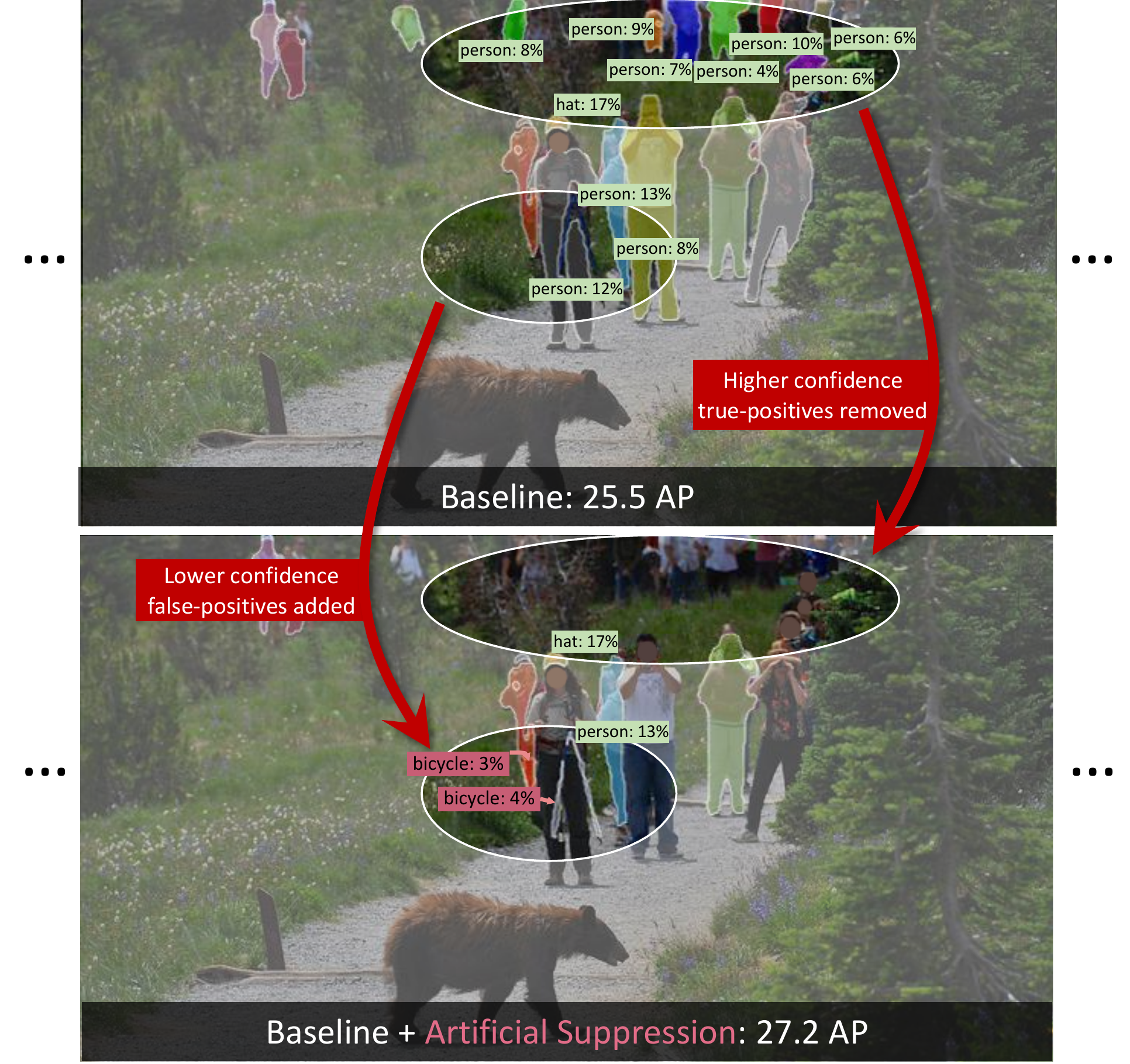}
    \caption{\textbf{The standard object detection average precision (AP) implementation can be gamed by an unintuitive re-ranking strategy.} Top: A detector normally outputs its top-$k$ most confident detections per image. Bottom: We discover an unintuitive re-ranking strategy that can increase AP substantially by reducing the number of detections output for frequent classes (\eg, `person') and increasing the number output for rarer classes (\eg, `bicycle'). This re-ranking balances AP better across categories, but counterintuitively removes some higher confidence true positives while also adding some lower confidence false positives, as shown above.
    We analyze why this happens in practice, how to fix it, and explore the consequences of the proposed solution.}
    \label{fig:teaser}
\end{figure}

\section{Introduction}

The task of object detection is commonly benchmarked by the mean of a per-category performance metric, usually average precision (AP)~\cite{everingham2010pascal,lin2014microsoft}.
This evaluation methodology is designed to treat all categories \emph{independently}: the AP for each category is determined by its confidence-ranked detections and is not influenced by the other categories.
On one-hand, this is a desirable property as it treats all classes equally.
On the other hand, it ignores cross-category score calibration, a key property in real-world use cases.

Surprisingly, in practice object detection benchmarking diverges from the goal of category-independent evaluation.
Cross-category interactions enter into evaluation due to a seemingly innocuous implementation detail: the number of detections per image, across all categories, is limited to make evaluation tractable~\cite{lin2014microsoft,gupta2019lvis}.
If a detector would exceed this limit, then a policy must be chosen to reduce its output.
The commonly used policy ranks all detections in an image by confidence and retains the top-scoring ones, up to the limit.
This policy naturally outputs the detections that are most likely to be correct according to the model.

However, this natural policy is not necessarily the best policy given the objective of maximizing AP\@.
We will demonstrate a counterintuitive result: there exists a policy, which can achieve higher AP, that discards a well-chosen set of higher-confidence detections in favor of promoting lower-confidence detections; see \Cref{fig:teaser}.
We first derive this result using a simple toy example with a perfectly calibrated detector.
Then, we show that given a real-world detection model, we can employ this new ranking policy to improve AP on the LVIS dataset~\cite{gupta2019lvis} by a non-trivial margin.
This policy is unnatural because it directly contradicts the model's confidence estimates---\emph{even when they are perfectly calibrated}---and shows that AP, as implemented in practice, can be vulnerable to gaming-by-re-ranking.

This analysis reveals that the default AP implementation neither achieves the goal of being independent per class nor, to the extent that it involves cross-category interactions, does it measure cross-category score calibration with a principled methodology.
Further, the metric can be gamed.
To address these limitations, first we fix AP to make it truly independent per class, and second, given the practical importance of calibration, we consider a complementary metric, \appool, that directly measures cross-category ranking.

Our fix to the standard AP implementation removes the detections-per-image limit and replaces it with a per-class limit over the entire evaluation set.
This simple modification leads to tractable, class-independent evaluation.
We examine how recent advances on LVIS fare under the new evaluation by benchmarking recently proposed loss functions, classifier head modifications, data sampling strategies, network backbones, and classifier retraining schemes.
Surprisingly, we find that many gains in AP stemming from these advances do not translate into improvements for the proposed category-independent AP evaluation.
This finding shows that the standard AP is sensitive to changes in cross-category ranking.
However, this sensitivity is an unintentional side-effect of the detection-per-image limit, not a principled measure of how well a model ranks detections across categories.

To enable more reliable benchmarking, we propose to \emph{directly} measure improvements to cross-category ranking with a complementary metric, \appool.
\appool pools detections from all classes and computes a single precision-recall curve --- the detection equivalent of micro-averaging from the information retrieval community~\cite{manning2008introduction}.
To optimize \appool, true positives for \emph{all} classes must rank ahead of false positives for \emph{any} class, making it a principled measure of cross-category ranking.
We extend simple score calibration approaches to work for large-vocabulary object detection and demonstrate significant \appool improvements that result in state-of-the art performance.

\section{Related work}\label{sec:related}

\smallsec{Large-vocabulary detection.}
Object detection research has largely focused on small-to-medium vocabularies (\eg, 20~\cite{everingham2010pascal} to 80~\cite{lin2014microsoft} classes), though notable exceptions exist~\cite{dean2013fast,hu2018learning}.
Recent detection benchmarks with hundreds~\cite{zhou2016semantic,kuznetsova2018open} to over one-thousand classes~\cite{gupta2019lvis} have renewed interest in large-vocabulary detection.
Most approaches re-purpose models originally designed for small vocabularies, with modifications aimed at class imbalance.
Over-sampling images with rare classes to mimic a balanced dataset~\cite{gupta2019lvis} is simple and effective.
Another strategy leverages advances from the long-tail \textit{classification} literature, including classifier retraining~\cite{kang2019decoupling,zhang2019study} and using a normalized classifier~\cite{liu2019large,wang2020seesaw}.
Finally, recent work proposes several new loss functions to reduce the penalty for predicting rare classes, \eg, equalization loss (EQL)~\cite{liu2019large}, balanced group softmax (BaGS)~\cite{li2020overcoming} or the CenterNet2 Federated loss~\cite{zhou2020cn2}.
We analyze these advances in large-vocabulary detection, finding that a number of them do not show improvements under our fixed, independent per-class AP evaluation, indicating that they improve existing AP by modifying cross-category rankings.

\smallsec{Detection evaluation.}
Average precision (AP) is the most common object detection metric, used by PASCAL~\cite{everingham2010pascal}, COCO~\cite{lin2014microsoft}, OpenImages~\cite{kuznetsova2018open}, and LVIS~\cite{gupta2019lvis}.
Conceptually, AP evaluates detectors independently for each class.
We show that common implementations deviate from this conceptual goal in important ways, and propose potential fixes and alternatives.
Prior work analyzing AP focuses on comparisons across classes, \eg Hoiem \etal~\cite{hoiem2012diagnosing} present a normalized average precision (AP$_N$) and Zhang \etal~\cite{zhang2019study} propose `sampled AP', but does not expose the issues covered in this paper.
Our procedural fix for AP computation removes the impact of cross-category scores on evaluation, and thus we propose a variant, \appool, which explicitly rewards better cross-category rankings.
From an information retrieval perspective, \appool is the micro-averaging counterpart to AP~\cite{manning2008introduction}, which evaluates macro-averaged performance, and has been used as a diagnostic in prior work~\cite{desai2011discriminative}.

\smallsec{Model calibration.}
A well-calibrated model is one that provides accurate probabilistic confidence estimates.
Calibration has been explored extensively in the classification setting, including parametric approaches, such as Platt scaling~\cite{platt1999probabilistic} and beta calibration~\cite{kull2017beta}, and non-parametric approaches, such as histogram binning~\cite{zadrozny2001obtaining}, isotonic regression~\cite{zadrozny2002transforming}, bayesian binning into quantiles (BBQ)~\cite{naeini2015obtaining}.
While small neural networks tend to be well-calibrated~\cite{niculescu2005predicting}, Guo \etal~\cite{guo2017calibration} show that deep networks are heavily uncalibrated.
Kuppers \etal~\cite{kuppers2020multivariate} extend this analysis to deep network based object detectors and show that size and position of predicted boxes helps reduce calibration error.
We also apply calibration strategies to object detectors, but find that \emph{per-class} calibration is crucial for improving \appool.

\section{Pitfalls of AP on large-vocabulary detection}\label{sec:rankings-and-ap}

Through both toy and real-world examples, we show that cross-category scores impact AP in counterintuitive ways.

\subsection{Background}\label{sec:setup-ap}
\vspace{-0.25em}
The standard object detection evaluation aims to evaluate each class independently.
In practice, however, this independence is broken due to an apparently harmless implementation detail:
to evaluate efficiently, benchmarks limit the number of detections a model can output per image (\eg to 100).
In practice, this limit is set (hopefully) to be high enough that detections beyond it are unlikely to be correct.
Importantly, this limit is shared across all classes, implicitly requiring models to rank predictions \textit{across} classes.

Our analysis shows that this detections-per-image limit, when used with a class-balanced evaluation like AP, can enable an unintuitive ranking policy to perform better than the natural policy of ranking detections by their estimated confidence.
The effect size is correlated with increasing the number of categories or the average instances per category.

\begin{figure}
    \subfloat[Left: Predictions from a \textit{perfectly calibrated} model: a prediction with confidence $s$ is correct $100 \cdot s\%$ of the time. Middle, Right: the two possible groundtruth scenarios and their probabilities.]{
        \raisebox{0.25\height}{\includegraphics[width=0.45\linewidth]{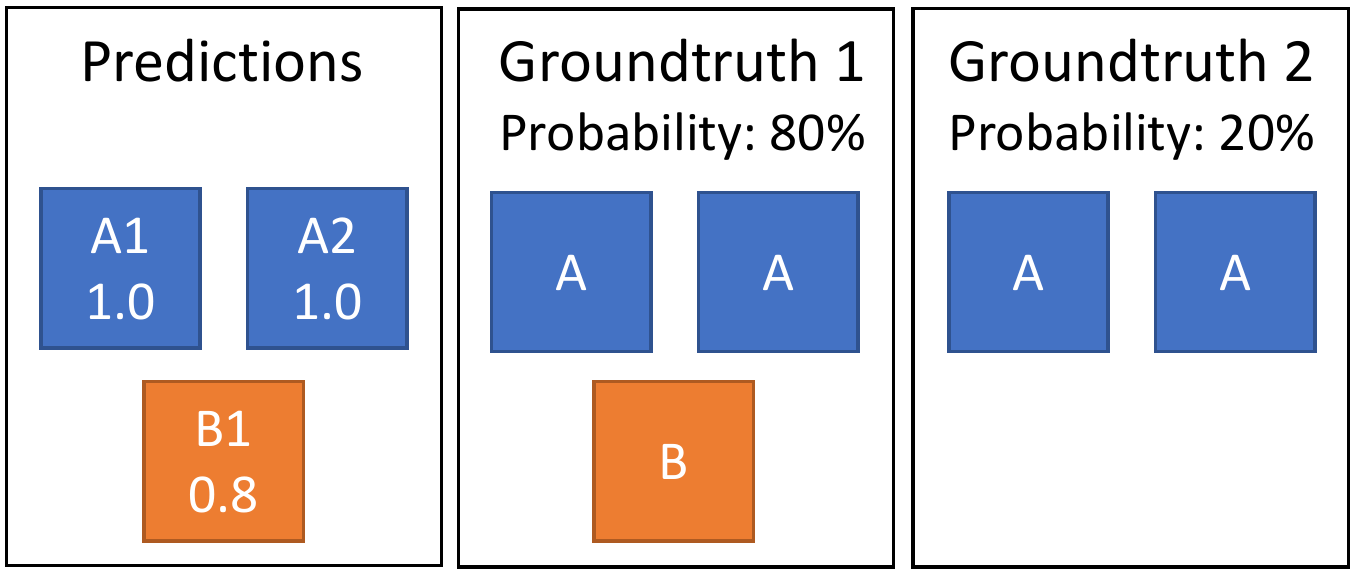}}
        \label{fig:balance_calib_a}
    }\hfill
    \subfloat[Two potential rankings of the predictions. With detections-per-image limited to 2, the two rankings report different predictions (\ie only those left of the dashed line).]{
        \includegraphics[width=0.45\linewidth]{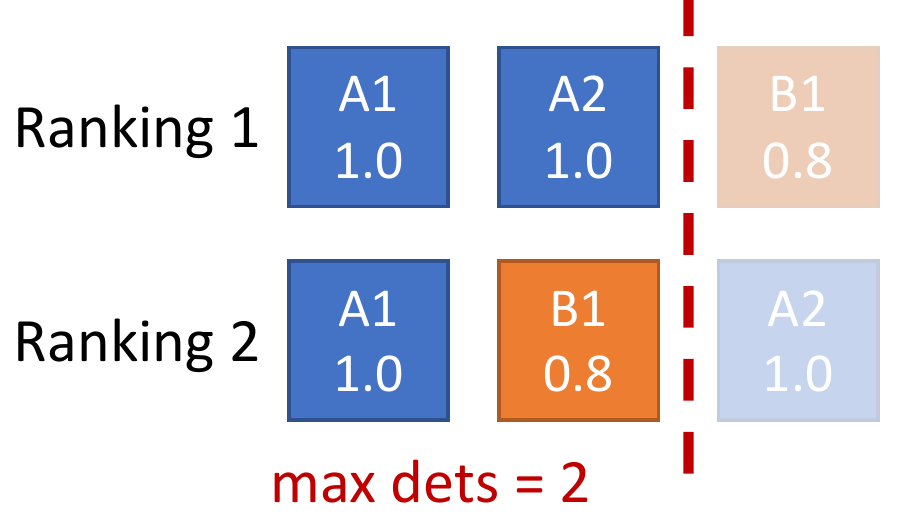}
        \label{fig:balance_calib_b}
    }\\[-1mm]
    \subfloat[Ranking 1 precision and recall. Since A1, A2 have a precision of 1.0, AP for class A is 1.0. Class B has no predictions, so the AP is 0.0, leading to an overall AP of 0.5.]{
        \includegraphics[width=0.46\linewidth]{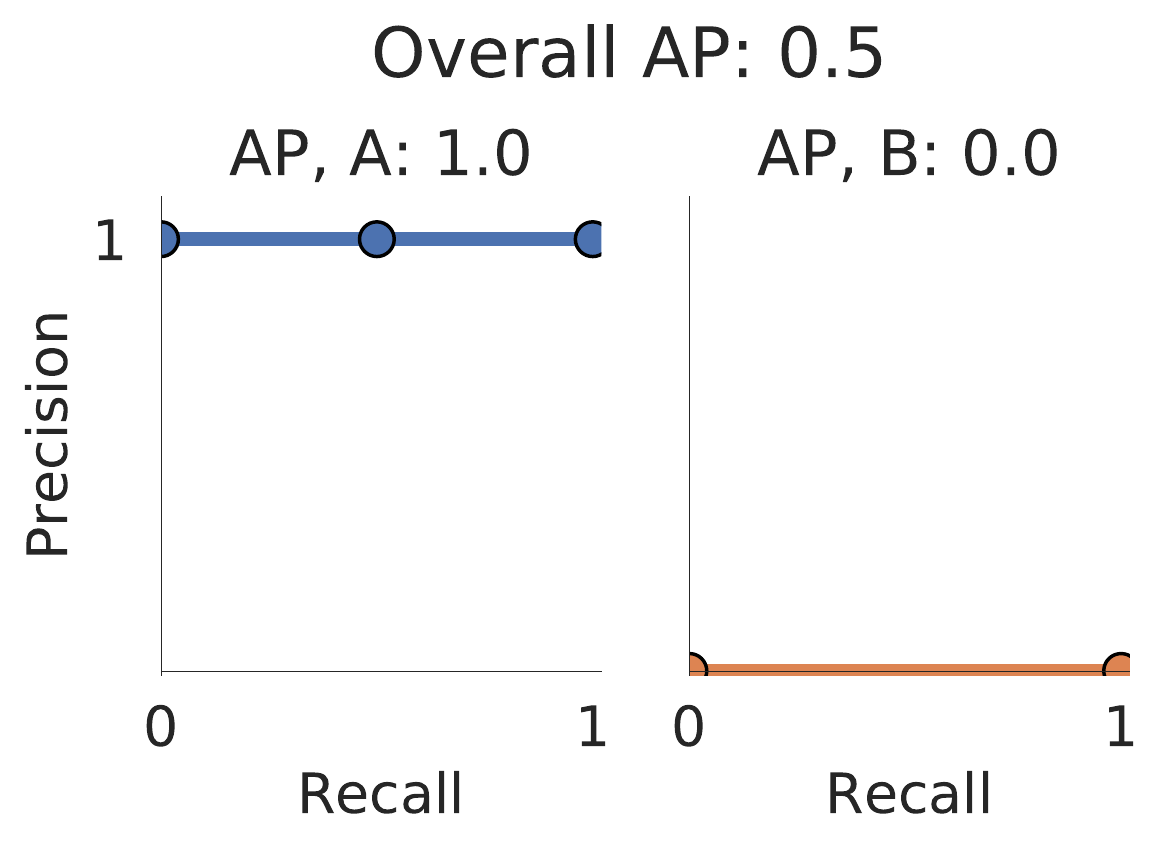}
        \label{fig:balance_calib_c}%
    }\hfill
    \subfloat[Ranking 2 precision and recall. AP for A is 0.5. For B: B1 is either a true positive (AP 1.0) or not (AP 0.0). On average, this results in AP 0.8 for B, and overall AP of 0.65.]{
        \includegraphics[width=0.46\linewidth]{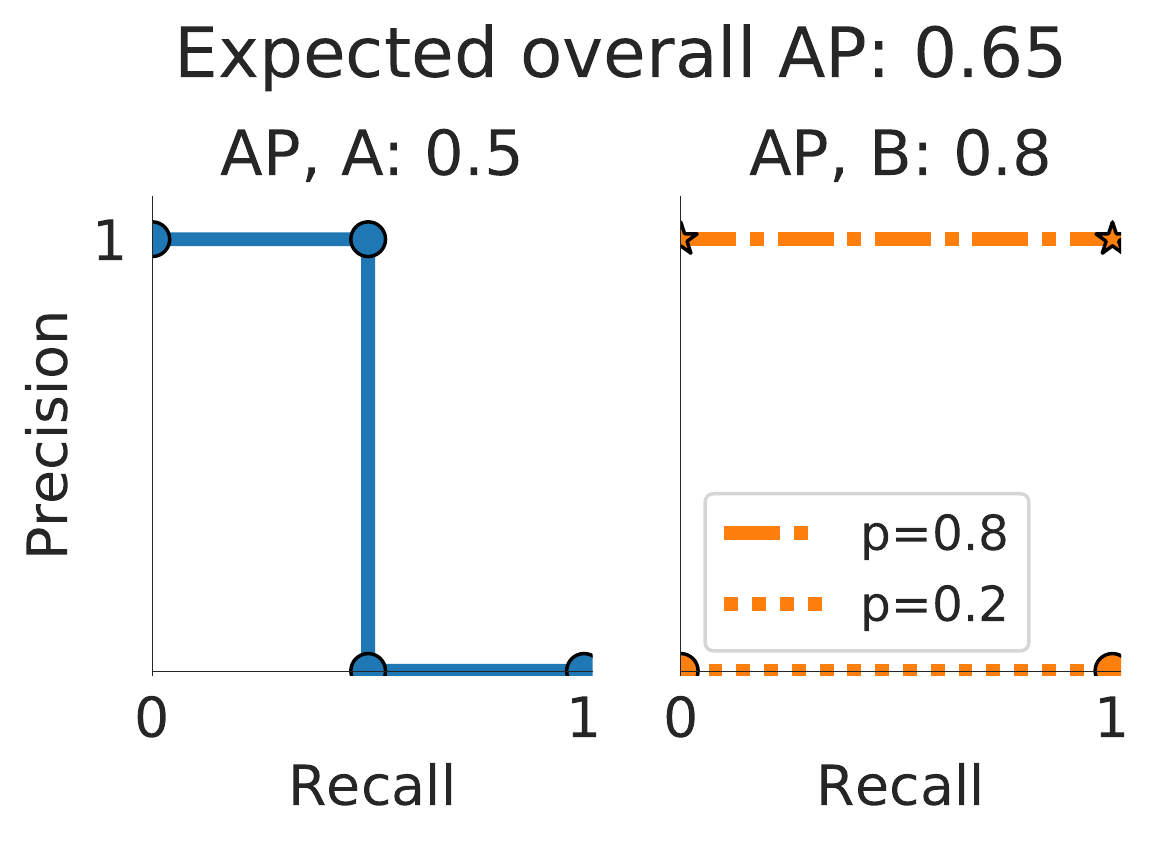}
        \label{fig:balance_calib_d}
    }
    \vspace{1.0ex}
    \caption{\textbf{Limiting detections-per-image rewards unintuitive rankings.} A toy scenario showing the interplay between a class-balanced AP evaluation and a limit on the number of detections per image. A perfectly calibrated model should output `Ranking 1' in (b) since it ranks detections that are more likely correct first. However, given a detections-per-image limit of 2, `Ranking 2' yields a higher AP even though it ranks a detection that is more likely incorrect (B1) ahead of one that is more likely correct (A2). Note that by removing the limit, the rankings across categories become fully independent and both rankings would result in an equal overall AP for the two rankings (0.75 in expectation; not visualized).
    \vspace{-2em}
    }
    \label{fig:balance_calib}
\end{figure}

\subsection{Analysis}

\vspace{-0.25em}
\smallsec{A toy example.}
Consider a toy evaluation on a dataset with two classes, as shown in~\Cref{fig:balance_calib}.
For simplicity, suppose we have access to a detector that is perfectly calibrated: when the model outputs a prediction with confidence $s$ (\eg 0.3), the prediction is a true positive $100 \cdot s\%$ (\eg 30\%) of the time.
We consider evaluating this model's outputs under two different rankings, using the standard class-balanced AP evaluation with a limit of two detections per image.

Under this setting, consider the predictions \wrt two possible groundtruth scenarios in \Cref{fig:balance_calib_a}.
The model predicts two instances for class A (A1, A2) with confidence $1.0$, and one instance for class B (B1), with confidence $0.8$.
Since the model is perfectly calibrated (by assumption), we know A1 and A2 are true positives $100\%$ of the time, while B1 is a true positive $80\%$ of the time.

With these predictions, consider the two potential rankings depicted in~\Cref{fig:balance_calib_b}.
Ranking 1 appears ideal: it ranks more confident detections before lower confident ones, as is standard practice.
By contrast, Ranking 2 is arbitrary: B1 is ranked above A2, despite having lower confidence.

Surprisingly, Ranking 2 \textit{outperforms} Ranking 1 under the AP metric with a limit of two detections per image, as shown in \Cref{fig:balance_calib_c} and \Cref{fig:balance_calib_d}.
While Ranking 1 gets a perfect AP of $1.0$ for class A, it gets $0$ AP for class B, leading to an overall AP of $0.5$.
By contrast, while Ranking 2 leads to a lower AP for class A ($0.5$), it scores an expected AP of $0.8$ for class B, yielding an overall AP of $0.65$!

Of course, this is a toy scenario, concocted to highlight an evaluation pitfall using an artificially low detections-per-image limit of only two predictions.
We now show that a similar effect exists for a real-world detection benchmark.

\smallsec{A real-world example.}
The LVIS~\cite{gupta2019lvis} dataset uses the evaluation described above, with a limit of 300 detections per image.
We investigate whether an artificial ranking policy, as in~\Cref{fig:balance_calib}, can lead to improved AP on this dataset.
Concretely, we evaluate a simple policy: we first discard all but the top $k$ scoring detections per class across the entire evaluation dataset.
Given the predictions in \Cref{fig:balance_calib_a}, applying this policy with $k=1$ leads to Ranking 2 from \Cref{fig:balance_calib_b}: an arbitrary ranking which, nevertheless, leads to a higher AP than the baseline Ranking 1.

This ranking policy, combined with the detections-per-image limit, is unintuitive: it explicitly discards high-scoring predictions for many classes in order to fit low-scoring predictions from other classes into the detections-per-image limit, as shown by our toy example in~\Cref{fig:balance_calib_b} and with real-world detections in~\Cref{fig:teaser}.
Using a baseline Mask R-CNN model~\cite{he2017mask} (see supp. for details), we find that this strategy, with $k=10,000$, improves LVIS AP by 1.2 points, and \apr by 2.9 points, as shown in \Cref{tab:artificially_limit_dets_per_class}.
Note that this results purely from a modified \textit{ranking policy}, without any changes to the evaluation or model.
This non-trivial improvement is roughly the magnitude achieved by a typical new method published at CVPR (\eg~\cite{tan2020equalization,li2020overcoming}).
The relatively larger improvement to \apr suggests that under the standard confidence-based ranking, accurate predictions for rare classes are crowded out by frequent class predictions due to the detections-per-image limit.

Although this limit appears high (at 300 detections-per-image), LVIS contains over a thousand object classes: even outputting a single prediction for each class is impossible under the limit.
The assumption is that detections beyond the first 300 are likely to be false positives.
\Cref{tab:higher_dets_per_im} verifies that this assumption is incorrect: increasing the limit on detections per image leads to significantly higher results on the LVIS dataset.
In particular, the AP for rare categories improves \textit{drastically} from 12.6 to 19.5 with a higher limit.

\begin{table}
    \tablestyle{3pt}{1.1}
    \begin{tabular}{rccccc}
        dets/class & dets/im & AP & \apr & \apc & \apf \\
        \midrule
        $\infty$ (Ranking 1)  & 300 & 22.6 \phantom{\bd{\dt{+1.2}}} & 12.6 \phantom{\bd{\dt{+2.9}}}  & 21.1 & 28.6 \\
        10,000 (Ranking 2)    & 300 & \bf 23.8 \bd{\dt{+1.2}} & 15.5 \bd{\dt{+2.9}} & 22.7 & 28.5 \\
    \end{tabular}
    \vspace{0.5ex}  %
    \caption{\textbf{Unintuitive Ranking 2 (\Cref{fig:balance_calib_b}) improves LVIS AP.} Artificially limiting the number of detections per class across the entire validation set leads to \textit{higher} LVIS AP when using the standard limit of 300 detections per image, perhaps paradoxically. In~\Cref{fig:teaser} we show how this ranking policy (which, again, improves AP) suppresses some higher-confidence detections in favor detections that the model estimates are more likely incorrect.}
    \label{tab:artificially_limit_dets_per_class}
\end{table}

\begin{table}
    \tablestyle{6pt}{1.1}
    \begin{tabular}{rcx{40}cc}
        & \multicolumn{4}{c}{LVIS} \\
        dets/\textbf{im} & AP & \apr & \apc & \apf \\
        \cmidrule(lr){1-1}
        \cmidrule(lr){2-5}
          100  & \gray{18.2} & \gray{6.5} & \gray{15.8} & \gray{26.1} \\
          300  & 22.6 & 12.6 & 21.1 & 28.6 \\
        1,000  & 25.0\td{+2.4} & 16.8\td{+4.2} & 24.1 & 29.7 \\
        2,000  & 25.6\td{+3.0} & 18.1\td{+5.5} & 24.6 & 29.9 \\
        5,000  & 26.0\td{+3.4} & 19.7\td{+7.1} & 24.9 & 30.0 \\
        10,000 & 26.1\tdb{+3.5} & 19.8\tdb{+7.2} & 25.0 & 30.1 \\
    \end{tabular}
    ~~
    \begin{tabular}{x{30}}
        COCO \\
        AP \\
        \cmidrule(lr){1-1}
        37.4 \\ %
        37.5\td{+0.1} \\ %
        37.5\td{+0.1} \\ %
        37.5\td{+0.1} \\ %
        37.5\td{+0.1} \\ %
        37.5\tdb{+0.1} \\ %
    \end{tabular}
    \vspace{0.5ex}  %
    \caption{\textbf{Increasing the limit on detections per image \textit{significantly} improves LVIS AP.} \apr improves by over 7 points (over 50\% relative), indicating many accurate rare class predictions are ignored due to the default limit of 300 detections per image. By contrast, this limit does not significantly impact COCO, which contains a significantly smaller vocabulary.\vspace{-1em}}
    \label{tab:higher_dets_per_im}
\end{table}

\smallsec{\change{When is gameability an issue?}}
Given the impact of the detections-per-image limit on LVIS, a natural question is whether this also affects the widely used COCO dataset.
\Cref{tab:higher_dets_per_im} shows that increasing this limit does not significantly change COCO AP, suggesting the limit has not negatively impacted COCO evaluation.
We hypothesize that this is due to the significantly smaller vocabulary in COCO relative to the detections limit: with only 80 classes, detections beyond the top 100 per image are unlikely to impact AP.

\begin{table}
\tablestyle{3pt}{1.1}
\begin{tabular}{ccccc@{\hskip 15px}ccc}
    & & & & \# instances & \multicolumn{2}{c}{AP@dets/im} & \\
    \multicolumn{3}{c}{Subset} & \# classes & per class & 300 & 5,000 & $\Delta$ \\
    \midrule
    R &   &   & 337   &   3.6 & 18.5 & 19.0 & +0.5 \\
      & C &   & 461   &  28.4 & 24.6 & 25.0 & +0.4 \\
      &   & F & 405   & 569.0 & 28.7 & 30.0 & +1.3 \\\midrule
    R & C &   & 798   &  17.9 & 22.2 & 23.3 & +1.1 \\
      & C & F & 866   & 281.2 & 24.7 & 27.4 & +2.7 \\
    R &   & F & 742   & 312.2 & 24.1 & 26.6 & +2.5 \\\midrule
    R & C & F & 1,203 & 203.4 & 22.6 & 26.0 & +3.4 \\
\end{tabular}
\vspace{0.5ex}  %
\caption{\textbf{Analyzing dets/image limit on LVIS subsets.} We restrict a baseline model to a subset of classes and evaluate on the subset.
`R', `C', and `F' indicate rare, common and frequent.
We compare the AP change ($\Delta$) at the default 300 dets/im limit \vs a high limit of 5,000. 
The change is more prominent for subsets with more classes and more instances per class, suggesting it is driven by both large vocabularies and the number of labeled objects.
}
\label{tab:lvis_subsets}
\end{table}

\change{To analyze this hypothesis, we evaluate on subsets of LVIS.
Given a baseline model trained on LVIS, we restrict its predictions to a subset of classes, and report AP with a low and a high detections-per-img limit in \Cref{tab:lvis_subsets}.
We find that on subsets which have small vocabularies and few labeled instances per class, the gap between AP in the two settings is small (0.4-0.5 points).
However, when there are many labeled instances in the evaluation set (as with the `F' subset), or the vocabulary is large (as with the second and third blocks of the table), the gap is much higher.
This suggests that AP is sensitive to the detections limit on large vocabulary datasets, particularly if they contain many labeled instances per image.
}

\begin{table}
    \tablestyle{3pt}{1.1}
    \begin{tabular}{rc@{\hspace{5ex}}c@{\hspace{7ex}}c@{\hspace{7ex}}cc}
        dets/\textbf{class} & dets/\textbf{im} & AP & \apr & \apc & \apf \\
        \midrule
        1,000  & $\infty$ & 21.9 & 17.7 & 22.2 & 23.5 \\
        5,000  & $\infty$ & 25.0\td{+3.1} & 19.5\td{+1.8} & 24.4 & 28.2 \\
        10,000 & $\infty$ & 25.6\td{+3.7} & 19.7\td{+2.0} & 24.7 & 29.1 \\
        30,000 & $\infty$ & 26.0\tdb{+4.1} & 19.8\td{+2.1} & 24.9 & 29.8 \\
        50,000 & $\infty$ & 26.0\tdb{+4.1} & 19.9\tdb{+2.2} & 25.0 & 30.0 \\
    \end{tabular}
    \vspace{0.5ex}  %
    \caption{\textbf{LVIS AP evaluation with varying limits on the number of detection/class, with no limit on detections/image.} A limit of 10,000 balances evaluation speed, memory, and AP well.}
    \label{tab:dets_per_cat}
\end{table}

\section{AP without cross-category dependence}
\achal{This entire section has been heavily edited/changed; for ease of reading, I haven't wrapped it in \texttt{\textbackslash change\{\}}.}

We now address this undesirable interaction between AP and cross-category scores.
We have already diagnosed that this interaction is caused by the detections-per-image limit.
In theory, then, the solution is simple: don't limit the number of detections per image.
Of course, this is impossible in practice, as we cannot evaluate infinite detections.
How, then, can we approximate this hypothetical evaluation?

\smallsec{Higher detections-per-image limit.}
A natural option is to have a large, but finite, detections-per-image limit.
Predictions beyond a very high limit are exceedingly unlikely to be correct, and thus may not affect the evaluation.
Indeed, \Cref{tab:higher_dets_per_im} shows that increasing the limit beyond 5,000 does not significantly affect AP.
Unfortunately, this results in prohibitively slower evaluation: on LVIS validation, a baseline model's outputs are $15\times$ larger using a limit of 5,000 detections than at the default limit of 300 (37GB \vs 2.4GB).
Moreover, submitting such results to an evaluation server, as required for the LVIS test sets, is impractical.

\smallsec{Limit detections-per-class.}
We now present an alternative, tractable implementation.
Rather than discarding low-scoring detections per image, we discard low-scoring detections per class across the dataset.
That is, given a model's output on the evaluation set, the benchmark would only evaluate the top $k$ predictions per class, discarding the rest.

We find that this strategy significantly reduces the storage and time requirements for evaluation.
\Cref{tab:dets_per_cat} shows that limiting detections to 10,000 per class across the dataset achieves a good balance.
This limit yields 98.5\% of full AP while increasing file size and evaluation time only by a factor of $2\times$ (compared to $15\times$ for the previous strategy), making evaluation tractable.
In principle, this limit depends on the size of the evaluation set, similar to how the standard per-image limit depends on the vocabulary size and labeling density.
In practice, the LVIS validation and test sets all contain 20,000 images and thus a single limit suffices.

This evaluation may appear similar to the undesirable Ranking 2 in \Cref{fig:balance_calib}.
However, Ranking 2 is an undesirable \textit{strategy} for resolving competition across classes, while our evaluation \textit{removes} this competition altogether by providing an independent detection budget per class.
This evaluation has a natural appeal when viewing detection as an information retrieval task, the field from which AP originates: the detector is allowed to `retrieve' up to $k$ detections (or `documents') per class from the entire evaluation set (or `corpus'). In practice, various strategies exist for efficiently selecting the top $k$ detections over a large set of images.

We recommend this latter strategy of limiting detections per class, with no limit per image.
In the remainder of the paper, we refer to the default evaluation (with a detections-per-image limit) as `\oldap' and our new, recommended version that limits detections per class as `\fixedap'.

\begin{table*}[t]
    \centering
    \subfloat[\textbf{Loss functions.} Choosing the right loss is more important under \oldap{}, providing an improvement of up to +2.4 AP\@.
    Under our proposed \fixedap{}, the impact of losses is reduced, to at most +0.8 AP.
    This result indicates that these loss functions may primarily modify cross-category rankings (also see~\Cref{fig:score-dist}).]{%
    \tablestyle{7pt}{1.1}
        \begin{tabular}[t]{r@{\hspace{3ex}}cx{40}}
     Loss & \oldap & \fixedap \\\midrule
     Softmax CE
             & 22.3 & 25.5 \\\midrule
     Sigmoid BCE
             & 22.5\td{+0.2} & 25.6\td{+0.1}\\
     EQL \cite{tan2020equalization}
             & 24.0\td{+1.7} & 26.1\td{+0.6}\\
     Federated \cite{zhou2020cn2}
             & \bf 24.7\td{+2.4} & \bf 26.3\td{+0.8}\\
     BaGS \cite{li2020overcoming}
             & 24.5\td{+2.2} & 25.8\td{+0.3}\\
    \phantom{foo} & & \\
    \phantom{foo} & & \\
    \phantom{foo} & & \\
    \end{tabular}
    \label{tab:ap_fixed_losses}
    }
    ~~~~
    \subfloat[\textbf{Classifier modifications.} We evaluate two ideas commonly used for improving long-tail detection: an objectness predictor (`Obj')~\cite{li2020overcoming}, and L2-normalizing both the linear classifier weights and input features (`Norm').
    Once again, we find that these components improve the baseline significantly under the \oldap,  but provide minor improvements under our \fixedap.
    Nevertheless, our results indicate these components provide a strong, simple baseline that erases the impact of the training loss choice.]{%
    \tablestyle{4pt}{1.1}
    \hspace{5ex}
    \begin{tabular}[t]{rcccx{50}}
     Loss         &
     Obj & Norm & \oldap & \fixedap \\\midrule
     \multirow{4}{*}{Softmax CE}
                & \gxmark & \gxmark & 22.3 & 25.5 \\
                & \cmark  & \gxmark & 23.2\td{+0.9} & 25.3\td{\textminus0.2} \\
                & \gxmark & \cmark  & 23.2\td{+0.9} & \bf 26.3\td{+0.8} \\
                & \cmark  & \cmark  & 24.4\td{+2.1} & 26.3\td{+0.8} \\\midrule
     Sigmoid BCE & \cmark & \cmark & 24.2\td{\tm0.2} & 26.3\td{+0.0} \\
     EQL \cite{tan2020equalization} & \cmark & \cmark & 24.7\td{+0.3} & 26.1\td{\textminus0.2} \\
     Federated \cite{zhou2020cn2} & \cmark & \cmark     & \bf 25.1\td{+0.7} & 26.3\td{+0.0}\\
     BaGS \cite{li2020overcoming} & \cmark & \cmark   & \bf 25.1\td{+0.7} & 26.2\td{\textminus0.1} \\
    \end{tabular}
    \label{tab:ap_fixed_components}
    \hspace{5ex}
    }
    \\
    \subfloat[\textbf{Samplers.}
    Category Aware Sampling (CAS) and Repeat Factor Sampling (RFS) are common sampling strategies for addressing class imbalance.
    While both strategies outperform the uniform sampling baseline under \oldap, only RFS provides significant improvements under \fixedap.
    ]{%
    \tablestyle{4pt}{1.1}
    \hspace{2ex}
    \begin{tabular}[t]{rcx{40}}
     Sampler   & \oldap & \fixedap \\\midrule
     Uniform   & 18.4 & 22.8 \\
     CAS       & 19.2\td{+0.8}     & 21.5\td{\tm 1.3}\\
     RFS       & \bf 22.3\td{+3.9} & \bf 25.5\td{+2.7} \\
     \\  %
    \end{tabular}
    \hspace{2ex}
    \label{tab:ap_fixed_samplers}
    }
    ~~~~
    \subfloat[\textbf{Classifier retraining.}
    We evaluate the efficacy of training detectors in two phases, a common technique~\cite{kang2019decoupling,wang2020seesaw}.
    Phase 1: the model is trained end-to-end with one sampler.
    Phase 2: only the final classification layer is trained, using a different sampler.
    This strategy improves \oldap, but not \fixedap, suggesting that classifier retraining may primarily modify cross-category rankings.]{%
    \tablestyle{4pt}{1.1}
    \begin{tabular}[t]{llcx{50}}
     Phase 1 & Phase 2 & \oldap & \fixedap \\\midrule
     RFS     & -       & 22.3              & \bf25.5 \\
     Uniform & RFS     & 21.6\td{\tm 0.7}  & 24.9\td{\tm 0.6} \\
     Uniform & CAS     & 23.1\td{+0.8}     & 24.9\td{\tm 0.6} \\
     RFS     & CAS     & \bf23.6\td{+1.3}  & \bf25.6\td{+0.1} \\
    \end{tabular}
    \label{tab:ap_fixed_crt}
    }
    ~~~~
    \subfloat[\textbf{Stronger backbones.}
    Using larger backbones consistently improves the detector under both \oldap{} and \fixedap, indicating, as one might expect, that larger backbones improve overall detection quality and not just cross-category rankings. ResNeXt-101 uses the 32x8d configuration.
]{%
    \tablestyle{4pt}{1.1}
    \begin{tabular}[t]{rcx{40}}
     Backbone   & \oldap & \fixedap \\\midrule
     ResNet-50           & 22.3              & 25.5 \\
     ResNet-101          & 24.6\td{+2.3}     & 27.7\td{+2.2} \\
     ResNeXt-101         & \bf 26.2\td{+3.9} & \bf 28.7\td{+3.2} \\
     \\
    \end{tabular}
    \hspace{2ex}
    \label{tab:ap_fixed_larger_models}
    }
    \vspace{1ex}
    \caption{\textbf{Impact of various design choices on the LVIS v1 validation dataset, comparing \oldap{} to \fixedap.} Unless specified otherwise, each experiment uses a ResNet-50 FPN Mask R-CNN model trained with Repeat Factor Sampling (RFS) for 180k iterations with 16 images per batch. All numbers are the average of three runs with different random seeds and initializations.}\vspace{-1mm}
    \label{tab:ap_fixed_big_table}
\end{table*}

\section{Impact on long-tailed detector advances}

We have shown that the current AP evaluation introduces subtle, undesirable interactions with cross-category rankings due to the detections-per-image limit.
However, it remains unclear to what extent this issue meaningfully affects prior conclusions drawn on LVIS.
To analyze this, we evaluate the importance of different design choices in LVIS detectors with the original evaluation (`\oldap'), with a limit of 300 predictions per image, and our modified evaluation (`\fixedap'), with a limit of 10,000 detections per class across the whole evaluation set (with no per-image limit).

\begin{figure}[t]
    \centering
    \includegraphics[width=0.9\linewidth]{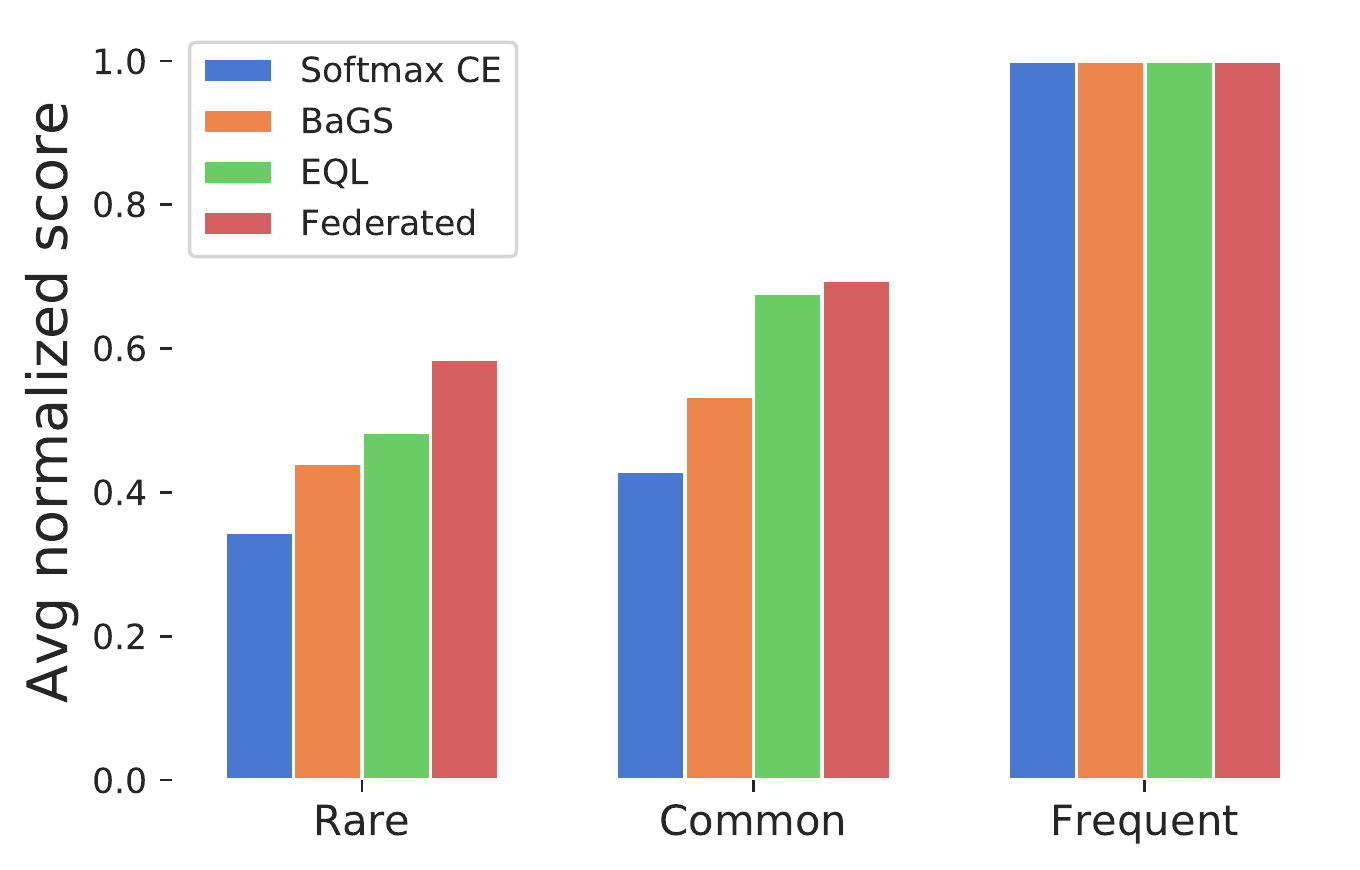}
    \caption{\textbf{Score distribution induced by different loss functions for LVIS rare, common, and frequent categories.} Compared to the baseline softmax CE loss, BaGS, EQL, and Federated losses tilt the distribution to be more uniform, modifying ranking of detections across categories.}\label{fig:score-dist}
    \vspace{-1em}
\end{figure}

\smallsec{Experimental setup.}
\label{sec:experimental_setup}
The following experiments use Mask R-CNN~\cite{he2017mask}.
Unless noted differently: we use a ResNet-50~\cite{he2016deep} backbone with FPN~\cite{lin2017feature} pre-trained on ImageNet~\cite{russakovsky2015imagenet} and fine-tuned on LVIS v1~\cite{gupta2019lvis} for 180k iterations with repeat factor sampling, minibatch of 16 images, learning rate of 0.02 decayed by 0.1$\times$ at 120k and 160k iterations, and weight decay of 1e\textminus 4. Batch norm~\cite{ioffe2015batch} parameters are frozen. Results are reported on LVIS v1 validation using the mean of three runs with different random seeds.

\subsection{Case studies}
\label{sec:case_studies}
\vspace{-0.5ex}
\smallsec{Loss functions.}
As discussed in \Cref{sec:related}, a number of new losses have been proposed in the past year.
We analyze three in particular: EQL~\cite{tan2020equalization}, BaGS~\cite{li2020overcoming}, and a `Federated' loss~\cite{zhou2020cn2}.
\Cref{tab:ap_fixed_losses} (first column) shows that, under the original evaluation, the choice of loss function can robustly improve the AP of a baseline model by up to 2.4 points, from 22.3 using softmax cross-entropy (CE) to 24.7 using the Federated loss.
These gains suggests the choice of loss function is important.
However, under our `\fixedap', the losses are more similar, differing by at most 0.8 points.

To gain insight into why the losses improve \oldap more than \fixedap, we plot the score distribution for the LVIS rare, common, and frequent categories (normalized so the average score for frequent categories is 1.0).
\Cref{fig:score-dist} shows that the EQL, BaGS, and Federated losses tilt the distribution to be more uniform relative to softmax CE loss.
This boosts the confidence of rare category detections, making them more likely to appear in the 300 detections-per-image limit.
This suggests that these losses change cross-category rankings compared to softmax CE loss in a way that \oldap rewards.
Because \fixedap is category independent, it does not reward cross-category ranking modifications.

\smallsec{Classifier heads.}
Next, we evaluate two common modifications to the linear classifier in detectors in~\Cref{tab:ap_fixed_components}.
The first modification trains a linear \textit{objectness} binary classifier in parallel to the $K$-way classifier~\cite{li2020overcoming,redmon2016you,wang2020seesaw}, denoted `Obj'.
The second L2-normalizes the input features and classifier weights during training and inference~\cite{wang2018cosface,liu2019large,wang2020seesaw}, denoted `Norm.'
We share implementation details in supplementary.

The first block in \Cref{tab:ap_fixed_components} shows that while adding an objectness predictor modestly improves \oldap (+0.9), it results in a slightly lower \fixedap (\textminus 0.2).
This discrepancy suggests the objectness predictor optimizes the ranking of predictions across classes, but doesn't meaningfully improve the quality of the detections.
On the other hand, using a normalized classifier consistently leads to higher accuracy under both \oldap{} (+0.9) and \fixedap (+0.8).
Finally, we find that applying both these modifications to the classifier results in a strong baseline under both \oldap{} and \fixedap.
The second block in \Cref{tab:ap_fixed_components} further shows that under \fixedap{}, the choice of loss function is largely irrelevant when both of these classifier modifications are used.

\smallsec{Sampling strategies.}
Modifying the image sampling strategy is a common approach for addressing class imbalance in LVIS.
\Cref{tab:ap_fixed_samplers} analyzes three strategies: Uniform, which samples images uniformly at random; Class Aware Sampling (CAS), which first samples a category and then an image containing that category; and Repeat Factor Sampling (RFS)~\cite{gupta2019lvis}, which oversamples images containing rare classes.
RFS consistently and significantly outperforms the others under both \oldap{} and \fixedap.
Surprisingly, while CAS outperforms uniform sampling under \oldap, it hurts accuracy under \fixedap, suggesting that CAS improves primarily due to how it ranks predictions across classes.

\smallsec{Classifier retraining.}
A common alternative to training with a single sampler is to train the model end-to-end using one sampler, and fine-tune the linear classifier with a different sampler~\cite{kang2019decoupling,zhang2019study}.
Under \oldap, carefully choosing the samplers for these phases appears important, improving by +1.3 AP.
However, under \fixedap, this improvement disappears, indicating that on LVIS, classifier retraining primarily improves by aligning scores across classes.

\smallsec{Stronger backbones.}
Finally, we evaluate the improvements due to stronger backbone architectures.
We evaluate four progressively stronger models: ResNet-50, ResNet-101~\cite{he2016deep}, and ResNeXt-101 32x8d~\cite{xie2017aggregated}. %
Unlike many other LVIS-specific design choices, we find that the choice of a larger backbone consistently improves accuracy for both \oldap{} and \fixedap.

\subsection{Discussion: something gained, something lost}
\fixedap makes AP evaluation category independent by design.
As a result, it is no longer vulnerable to gaming-by-re-ranking, as we demonstrate is possible with \oldap in \Cref{sec:rankings-and-ap}.
\change{However, by benchmarking several recent advances in long-tailed object detection we observe evidence that several of the improvements may be due to better cross-category rankings, because the improvements that were observed with \oldap largely disappear when evaluated with \fixedap.
While \oldap improperly evaluated calibration, \fixedap is \textit{invariant} to calibration: \ie, per-category, monotonic score transformations do not change \fixedap.}

\change{Neither \oldap nor \fixedap appropriately specifies how detectors should be deployed in the real world, a task which \textit{requires} score calibration.
In the simplest example, one may want to produce a demo that visualizes all detections above a global score threshold (\eg 0.5) and expect to see consistent results across all categories.}
Given this practical demand, we consider in the next section a variant of AP, called \appool, that directly rewards cross-category rankings, without the vulnerability to gaming displayed by \oldap.
Furthermore, we develop a simple detector score calibration method and show that it improves \appool.

\begin{table}
    \tablestyle{8pt}{1.1}
    \begin{tabular}{rcx{35}cc}
                & \multicolumn{4}{c}{\appool} \\
        dets/im & AP & \apr & \apc & \apf \\
        \midrule
        300    & 26.2 & \phantom{1}8.0 & 16.7 & 27.0  \\
        1,000  & 26.8\td{+0.6} & 10.6\td{+2.6} & 19.8 & 27.6 \\
        2,000  & 27.0\tdb{+0.8} & 11.0\td{+3.0} & 20.5 & 27.7 \\
        5,000  & 27.0\tdb{+0.8} & 11.3\tdb{+3.3} & 20.8 & 27.7 \\
        10,000 & 27.0\tdb{+0.8} & 11.3\tdb{+3.3} & 20.8 & 27.7 \\
    \end{tabular}
    \vspace{0.5ex}  %
    \caption{\textbf{Impact of limiting detections-per-image on \appool.} As expected, \appool is less sensitive to this limit than \oldap because each instance, rather than each class, is weighted equally.}
    \label{tab:higher_dets_per_im_with_pool}
\end{table}

\section{Evaluating cross-category rankings}

An independent, per-class evaluation is appealing in its simplicity.
Most practical applications, however, require comparing the confidence of predictions across classes to form a unified understanding of the objects in an image.
As an extreme example, note that a detector can output arbitrary range of scores for each class for a truly independent evaluation: that is, all detections for one class (say, `banana') may have confidences above $0.5$, while all detections for another class (say, `person') have confidences below $0.5$.
Using such a detector in practice requires carefully calibrating scores across classes---an open challenge that is not evaluated by current detection evaluations.

\subsection{\appool: A cross-category rank sensitive AP}
To address this, we consider a complementary metric, \appool, which explicitly evaluates detections across all classes together~\cite{desai2011discriminative}.
To do this, we first match predictions to groundtruth per-class, following the standard evaluation.
Next, instead of computing a precision-recall (PR) curve for each class, we pool detections across all classes to generate a single PR curve across all classes, and compute the Average Precision on this curve to get \appool.

This evaluation has two key properties.
First, it ranks detections across all classes to generate a single precision-recall curve, incentivizing detectors to rank confident predictions above lower confidence ones.
Second, it weights all groundtruth instances, rather than classes, equally.
This removes a counterintuitive effect, illustrated in~\Cref{fig:balance_calib}, that can occur with class averaging.
Further, it reduces the impact of the detections-per-image limit, as low-confidence predictions for some rare classes do not significantly impact the evaluation.
Because of this, however, the evaluation is influenced more by frequent classes than rare ones.
To analyze performance for rare classes, we further report three diagnostic evaluations which evaluate predictions only for classes within a specified frequency: \appoolr{} (for rare classes), \appoolc{} (common), and \appoolf{} (frequent).

\subsection{Analysis}

\begin{table}[t]
    \tablestyle{4pt}{1.1}
    \begin{tabular}{rx{12}@{\hskip 6ex}x{12}x{12}x{12}x{12}@{\hskip 6ex}x{12}x{12}x{12}}
        & \multicolumn{4}{c}{\fixedap} & \multicolumn{4}{c}{\appool} \\
        \cmidrule(lr){2-5}\cmidrule(lr){6-9}
     Loss        & AP   & \apr & \apc & \apf & AP   & \apr & \apc & \apf \\\midrule
     Softmax CE  &
      25.5          & 18.9 & 24.9 & 29.1 &     25.6          & 11.5 & 20.5 & 26.2 \\
     Sigmoid BCE &
      25.6\td{+0.1} & 19.4 & 24.9 & 28.9 &     25.6\td{+0.0} & 10.8 & 20.1 & 26.1 \\
     EQL \cite{tan2020equalization} &
      26.1\td{+0.6} & 19.9 & 26.1 & 28.9 &     25.9\td{+0.3} & 11.3 & 22.9 & 26.3 \\
     Federated \cite{zhou2020cn2} &
  \bf 26.3\td{+0.8} & 20.7 & 24.9 & 30.2 & \bf 27.8\td{+2.2} & 16.1 & 22.0 & 28.2 \\
     BaGS \cite{li2020overcoming} &
      25.8\td{+0.3} & 17.9 & 25.6 & 29.5 &     26.0\td{+0.4} & \phantom{0}9.1 & 20.8 & 26.4 \\
    \end{tabular}
    \vspace{0.5ex}  %
    \caption{\textbf{\fixedap and \appool{} for models trained with varying losses.} Federated significantly outperforms others under \appool.}
    \label{tab:ap_pool_losses}
\end{table}

\smallsec{How does the dets/im limit affect \appool?}
\Cref{tab:higher_dets_per_im_with_pool} analyzes how the detections-per-image limit impacts \appool.
As expected, increasing this limit does not significantly affect \appool{}: while AP can change drastically due to a few additional true positives for rare classes, \appool{} treats true positives for all classes equally.
Increasing the limit beyond $300$ detections improves the diagnostic \appoolr{} metric, but only mildly improves \appool{} by 0.8 points.
Nonetheless, for consistency, we evaluate models with the same detections as \fixedap: the top 10,000 per class, with no per-image limit. %

\smallsec{Do losses impact \appool?}
Next, we analyze various losses under \appool{}, though we also analyze other detector components in supp.
\Cref{tab:ap_pool_losses} compares losses under \fixedap and \appool.
Perhaps surprisingly, while EQL and BaGS do not meaningfully impact \appool, the Federated loss improves by 2.2 points over the baseline softmax CE loss.
This provides a new perspective for the Federated loss:
Although it does not explicitly calibrate models, it improves \textit{cross-category} ranking of predictions compared to other losses.

\subsection{Calibration}

We now propose a simple and effective strategy for improving \appool.
We re-purpose classic techniques for calibrating model uncertainty for the task of large-vocabulary object detection.
Calibration aims to ensure that the model's confidence for a prediction corresponds to the probability that the prediction is correct.
In the detection setting, if a model detects a box with confidence $s$, it should correctly localize a groundtruth box of the same category $s\%$ of the time~\cite{kuppers2020multivariate}.
While this property is not necessary for \appool, it provides a sufficient condition for improving cross-category rankings (\appool only requires that true positives are ranked higher than false positives across all classes, without requiring the scores to be \emph{probabilistically} calibrated).

\begin{figure}
    \includegraphics[width=\linewidth,trim=0 190px 0 0,clip]{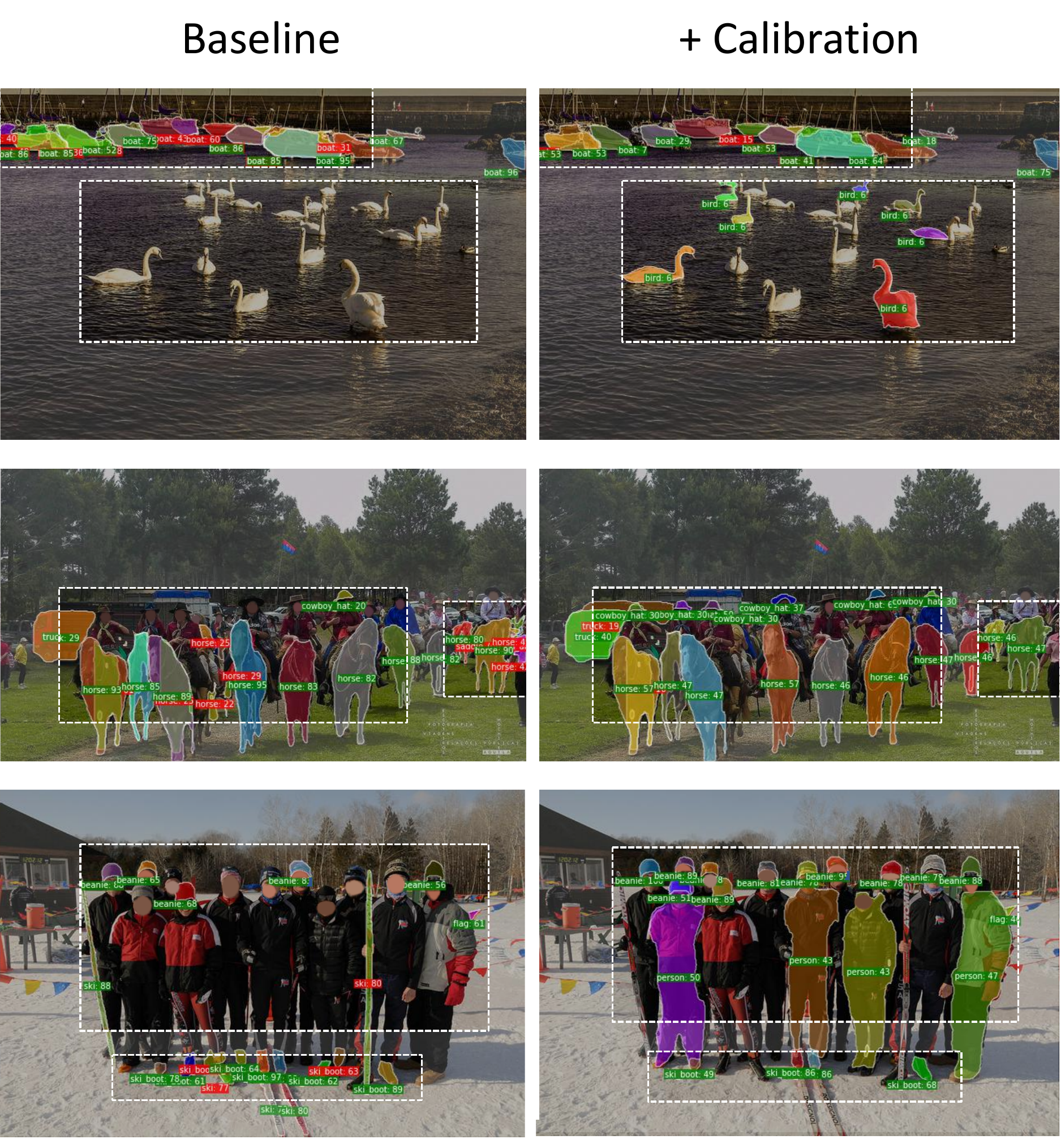}
    \caption{Examples illustrating the effect of calibration.
    Each row shows the 20 highest-scoring predictions from the baseline, uncalibrated model (left) and its calibrated version (right).
    True-positives and false-positives (at IoU 0.5) are indicated with a green and red label, respectively.
    The calibrated model increases the rank of low-confidence but accurate predictions, such as the `bird's (top row) and `cowboy hat's (bottom),
    over incorrect predictions with artificially high scores, such as some `boat's (top), and `horse's (bottom).}
    \label{fig:calibrated_qualitative}
\end{figure}
Following~\cite{kuppers2020multivariate}, we analyze various calibration strategies: histogram binning~\cite{zadrozny2001obtaining}, Bayesian Binning into Quantiles (BBQ)~\cite{naeini2015obtaining}, beta calibration~\cite{kull2017beta}, isotonic regression~\cite{zadrozny2002transforming}, and Platt scaling~\cite{platt1999probabilistic}.
Prior work on calibrating detectors applies calibration strategies to predictions across all classes~\cite{kuppers2020multivariate}.
However, this approach does not account for class frequency: rare classes may, for example, have lower-scoring predictions than frequent classes.
Instead, we propose to calibrate each class individually, allowing the method to boost scores of under-confident classes and diminish scores of over-confident classes.

Standard calibration strategies require a held-out dataset for calibration.
However, in the large-vocabulary setting, many classes have only a handful of examples in the entire dataset.
We instead calibrate directly on the \textit{training} set.
To understand the impact of this choice, we also report an upper-bound by calibrating on the \textit{validation} set.

\Cref{tab:ap_pool_calib} reports \appool{} using various calibration approaches applied to a model trained with the Federated loss.
The results show that
calibrating per class improves \appool{} by 1.7 points, from 27.8 to 29.5, and the choice of calibration strategy is not critical.
Surprisingly, calibrating on the \textit{validation} set, as in the second block, outperforms training set calibration by only 0.8 points, suggesting
that calibrating on the training set is a viable strategy.
However, calibrating on the validation set significantly improves \appoolr while calibrating on the training set \textit{harms} \appoolr, indicating that calibrating rare classes remains an open challenge.
\Cref{fig:calibrated_qualitative} presents qualitative examples of this improvement: calibration increases the scores of underconfident, accurate predictions from some classes (\eg `cowboy hat') and suppresses overconfident predictions from others (\eg `horse').

\begin{table}[t]
    \tablestyle{8pt}{1.1}
    \begin{tabular}{rcx{35}cc}
        Calibration      & \appool  & \appoolr  & \appoolc & \appoolf \\\midrule
        Uncalibrated     &     27.8          & \bf 16.1 &     22.0 &     28.2 \\
        Histogram Bin    &     28.6\td{+0.8} &     12.4 &     20.6 &     29.2 \\
        BBQ (AIC)        &     28.8\td{+1.0} &     13.6 &     21.6 &     29.3 \\
        Beta calibration & \bf 29.5\td{+1.7} &     12.8 &     22.7 & \bf 30.0 \\
        Isotonic reg.    &     28.3\td{+0.5} &     14.4 &     22.2 &     28.7 \\
        Platt scaling    & \bf 29.5\td{+1.7} &     13.1 & \bf 22.8 & \bf 30.0 \\
     \hline
     \multicolumn{5}{c}{
       \cellcolor{gray!15}Calibrate on validation (upper-bound oracle)
     } \\
     \hline
     HistBin            &     30.1\td{+2.3} &     24.4 & \bf 27.8 &     30.2 \\
     BBQ (AIC)          &     30.0\td{+2.2} &     22.9 &     26.9 &     30.2 \\
     Beta calibration   &     29.8\td{+2.0} &     22.4 &     25.2 &     30.1 \\
     Isotonic reg.      & \bf 30.3\td{+2.5} & \bf 24.6 &     27.2 & \bf 30.4 \\
     Platt scaling      &     29.8\td{+2.0} &     22.2 &     24.9 &     30.1 \\
    \end{tabular}
    \vspace{0.5ex}  %
    \caption{\textbf{Calibrating detection outputs on the train set significantly improves AP pooled.} The gains are due to improved rankings across categories. Calibrating on validation significantly improves \appoolr, indicating calibration remains challenging in the tail.
     All models trained with the Federated loss.}
    \label{tab:ap_pool_calib}
\end{table}

\section{Discussion}

Robust, reliable evaluations are critical for advances in large-vocabulary detection.
Our analysis reveals that current evaluations fail to properly handle cross-category interactions by neither eliminating them (as intended) nor evaluating them in a principled fashion (as potentially desired).
We show that, as a result, the current AP implementation (\oldap) is vulnerable to gaming.
We propose \fixedap, which addresses this gameability by removing the effect of cross-category score calibration, and recommend it as a \textit{replacement} for \oldap moving forward.
\fixedap provides new conclusions about the importance of different LVIS advances.
Finally, we recommend a complementary \textit{diagnostic} metric, \appool, for applications requiring cross-category score calibration, and show that a simple calibration strategy offers off-the-self detectors solid improvements to \appool.

\newpage

\begin{appendices}
\begin{center}{\bf \Large Appendix}\end{center}
We first present additional analysis of our experiments in \Cref{sec:supp_analysis}.
\Cref{sec:supp_analysis_pooled} reports \appool for all experiments in \Cref{tab:ap_fixed_big_table}, and discusses key results.
\Cref{sec:supp_analysis_crt} analyzes all variants of classifier retraining.
\Cref{sec:supp_analysis_stronger} analyzes losses and classifier modifications using a stronger baseline detector.
\Cref{sec:supp_analysis_nms} presents the impact of non-max suppression on \oldap compared to \fixedap.
Finally, \Cref{sec:supp_implementation} presents implementation details, including the experimental setup for tables in the main paper, classifier modifications, and the RegNetY-4GF model used in this appendix.

\section{Additional analyses}
\label{sec:supp_analysis}

\subsection{\appool: Long-tail detector advances}
\label{sec:supp_analysis_pooled}

\Cref{tab:ap_fixed_pool_all} reports \appool for all experiments in \Cref{tab:ap_fixed_big_table}.
We highlight a few results here.
\Cref{tab:ap_fixed_pool_losses} reports the same results as \Cref{tab:ap_pool_losses}:
while most losses do not improve \appool, the Federated loss provides significant improvements of +2.2 points under \appool, indicating it helps to calibrate cross-category scores.
\Cref{tab:ap_fixed_pool_components} shows that the classifier modifications described in \Cref{sec:case_studies} provide small improvements in \appool (+0.7 using both modifications).
Surprisingly, the Federated loss performs slightly better without these modifications under \appool (27.8 without modifications in \Cref{tab:ap_fixed_pool_losses} \vs 27.6 with modifications).
Even with both classifier modifications, the choice of loss is important under \appool, as the softmax CE loss significantly underperforms the Federated loss.
\Cref{tab:ap_fixed_pool_samplers} shows the uniform sampler performs on par with Repeat Factor Sampling, likely because \appool weights all instances equally, while \fixedap and \oldap weight each \textit{class} equally.
\Cref{tab:ap_fixed_pool_crt} shows that two-stage training does not improve \fixedap or \appool significantly.
Finally, larger backbones consistently improve \oldap, \fixedap, and \appool (\Cref{tab:ap_fixed_pool_larger_models}).

\subsection{Classifier retraining}
\label{sec:supp_analysis_crt}
\Cref{tab:ap_fixed_crt} evaluates the efficacy of training detectors in two phases, using a few different data sampling configurations.
For completeness, we report results using all sampler configurations in \Cref{tab:supp_ap_fixed_pool_crt_all}.
These results further support the conclusions from the main paper: while certain configurations improves over the baseline under \oldap (up to +1.3AP), they provide little to no improvements under \fixedap, suggesting they modify cross-category rankings.
However, they also do not impact \appool, indicating they do not meaningfully improve the calibration of scores across categories --- \ie, they may take advantage of the vulnerability in \oldap discussed in \Cref{sec:rankings-and-ap}.

\begin{table*}[t]
    \centering
    \subfloat[\textbf{Loss functions.} Most losses perform equally under \appool, with the exception of Federated loss, which significantly improves \appool by 2.2 points.]{%
    \tablestyle{7pt}{1.1}
        \begin{tabular}[t]{r@{\hskip 10pt}c@{\hskip 15pt}c@{\hskip 5pt}x{40}}
     Loss & \oldap & \fixedap & \appool \\\midrule
     Softmax CE   &     22.3          &     25.5          &     25.6 \\\midrule
     Sigmoid BCE  &     22.5\td{+0.2} &     25.6\td{+0.1} &     25.6\td{+0.0} \\
     EQL \cite{tan2020equalization}
                  &     24.0\td{+1.7} &     26.1\td{+0.6} &     25.9\td{+0.3} \\
     Federated \cite{zhou2020cn2}
                  & \bf 24.7\td{+2.4} & \bf 26.3\td{+0.8} & \bf 27.8\td{+2.2} \\
     BaGS \cite{li2020overcoming}
                  &     24.5\td{+2.2} &     25.8\td{+0.3} &     26.0\td{+0.4} \\
    \phantom{foo} & & \\
    \phantom{foo} & & \\
    \phantom{foo} & & \\
    \end{tabular}
    \label{tab:ap_fixed_pool_losses}
    }
    ~~~~
    \subfloat[\textbf{Classifier modifications.} We evaluate two ideas commonly used for improving long-tail detection: an objectness predictor (`Obj')~\cite{li2020overcoming}, and L2-normalizing both the linear classifier weights and input features (`Norm').
    These components mildly improve \appool, and the Federated loss still outperforms all other losses.
    Perhaps surprisingly, the modifications hurt the Federated loss (27.6 with \vs 27.8 without in \Cref{tab:ap_fixed_pool_losses}).]{%
    \tablestyle{4pt}{1.1}
    \begin{tabular}[t]{rccc@{\hskip 15pt}c@{\hskip 5pt}x{50}}
     Loss         &
     Obj & Norm & \oldap & \fixedap & \appool \\\midrule
     \multirow{4}{*}{Softmax CE}
                & \gxmark & \gxmark & 22.3              & 25.5              & 25.6 \\
                & \cmark  & \gxmark & 23.2\td{+0.8}     & 25.3\td{\tm0.2}   & 26.2\td{+0.6} \\
                & \gxmark & \cmark  & 23.2\td{+0.8}     & \bf 26.3\td{+0.8} & 25.7\td{+0.1} \\
                & \cmark  & \cmark  & \bf 24.4\td{+2.0} & \bf 26.3\td{+0.8} & \bf 26.3\td{+0.7} \\\midrule
     Sigmoid BCE & \cmark & \cmark  & 24.2\td{\tm0.2}   & \bf 26.3\td{+0.0} & 26.6\td{+0.3} \\
     EQL \cite{tan2020equalization}
                & \cmark & \cmark   & 24.7\td{+0.3}     & 26.1\td{\tm0.2}   & 26.6\td{+0.3} \\
     Federated \cite{zhou2020cn2}
                & \cmark & \cmark   & \bf 25.1\td{+0.9} & \bf 26.3\td{+0.0} & \bf27.6\td{+1.3} \\
     BaGS \cite{li2020overcoming}
                & \cmark & \cmark   & \bf 25.1\td{+0.9} & 26.2\td{\tm0.1}   & 25.9\td{\tm0.4} \\
    \end{tabular}
    \label{tab:ap_fixed_pool_components}
    }
    \\
    \subfloat[\textbf{Samplers.}
    Category Aware Sampling (CAS) and Repeat Factor Sampling (RFS) are common sampling strategies for addressing class imbalance.
    As \appool weights all instances equally, unlike \fixedap and \oldap (which weight all classes equally), uniform sampling performs on par with RFS.
    ]{%
    \tablestyle{1.5pt}{1.1}
    \begin{tabular}[t]{rc@{\hskip 12pt}c@{\hskip 2pt}x{45}}
     Sampler   & \oldap & \fixedap & \appool \\\midrule
     Uniform   &     18.4          &     22.8            & \bf25.7 \\
     CAS       &     19.2\td{+0.8} &     21.5\td{\tm1.3} &    21.7\td{\tm 4.0} \\
     RFS       & \bf 22.3\td{+3.9} & \bf 25.5\td{+2.7}   &    25.6\td{\tm 0.1} \\
     \\  %
    \end{tabular}
    \label{tab:ap_fixed_pool_samplers}
    }
    ~~~~
    \subfloat[\textbf{Classifier retraining.}
    We evaluate the efficacy of training detectors in two phases, a commonly used technique~\cite{kang2019decoupling,wang2020seesaw}.
    In Phase 1, the model is trained end-to-end with one sampler.
    In Phase 2, only the final classification layer is trained, using a different sampler.
    This strategy provides only minor improvements on \appool.]{%
    \tablestyle{1.5pt}{1.1}
    \begin{tabular}[t]{llc@{\hspace{15pt}}cx{40}}
     Phase 1 & Phase 2 & \oldap & \fixedap & \appool \\\midrule
     RFS     & -       &     22.3            &     25.5            &     25.6            \\
     Uniform & RFS     &     21.6\td{\tm0.7} &     24.9\td{\tm0.6} & \bf 25.8\td{+0.2}   \\
     Uniform & CAS     &     23.1\td{+0.8}   &     24.9\td{\tm0.6} &     25.6\td{+0.0}   \\
     RFS     & CAS     & \bf 23.6\td{+1.3}   & \bf 25.6\td{+0.1}   &     25.5\td{\tm0.1} \\
    \end{tabular}
    \label{tab:ap_fixed_pool_crt}
    }
    ~~~~
    \subfloat[\textbf{Stronger backbones.}
    Using larger backbones consistently improves the detector under \oldap{}, \fixedap, and \appool, indicating, as one might expect, that larger backbones improve overall detection quality and not just cross-category rankings. ResNeXt-101 uses the 32$\times$8d configuration.]{%
    \tablestyle{1.5pt}{1.1}
    \begin{tabular}[t]{rc@{\hspace{15pt}}cx{40}}
     Backbone   & \oldap & \fixedap & \appool \\\midrule
     ResNet-50           &     22.3          &     25.5          &     25.6          \\
     ResNet-101          &     24.6\td{+2.3} &     27.7\td{+2.2} &     27.2\td{+1.6} \\
     ResNeXt-101         &     \bf 26.2\td{+3.9} &     \bf 28.7\td{+3.2} &     \bf 29.0\td{+3.4} \\
     \\
    \end{tabular}
    \label{tab:ap_fixed_pool_larger_models}
    }
    \vspace{1ex}
    \caption{\textbf{Impact of various design choices on the LVIS v1 validation dataset, comparing \oldap{}, \fixedap and \appool.} We report all experiments from Table 4 of our main paper using \appool in addition to \fixedap and \oldap. As in the main paper, unless specified otherwise, each experiment uses a ResNet-50 FPN Mask R-CNN model trained with Repeat Factor Sampling (RFS) for 180k iterations with 16 images per batch. All numbers are the average of three runs with different random seeds and initializations.}
    \label{tab:ap_fixed_pool_all}
\end{table*}

\begin{table}
    \tablestyle{7pt}{1.1}
    \begin{tabular}[t]{llcc@{\hskip 7pt}x{40}}
     Phase 1 & Phase 2 & \oldap & \fixedap & \appool \\\midrule
     RFS     & -       & 22.3              & 25.5             & 25.6 \\\midrule
     Uniform & Uniform & 19.3\td{\tm 3.0}  & 24.0\td{\tm 1.5} & \bf 25.8\td{+0.2} \\
     Uniform & RFS     & 21.6\td{\tm 0.7}  & 24.9\td{\tm 0.6} & \bf 25.8\td{+0.2} \\
     Uniform & CAS     & 23.1\td{+0.8}     & 24.9\td{\tm 0.6} & 25.6\td{+0.0} \\\midrule
     RFS     & Uniform & 20.8\td{\tm 1.5}  & 25.4\td{\tm 0.1} & 25.8\td{+0.2} \\
     RFS     & RFS     & 22.6\td{+0.3}     & \bf25.8\td{+0.3} & 25.7\td{+0.1} \\
     RFS     & CAS     & \bf23.6\td{+1.3}  & 25.6\td{+0.1}    & 25.5\td{\tm 0.1} \\\midrule
     CAS     & Uniform & 17.4\td{\tm 4.9}  & 21.1\td{\tm 4.4} & 22.1\td{\tm 3.5} \\
     CAS     & RFS     & 18.2\td{\tm 4.1}  & 21.3\td{\tm 4.2}    & 22.1\td{\tm 3.5}\\
     CAS     & CAS     & 19.1\td{\tm 3.2}  & 21.2\td{\tm 4.3} & 21.7\td{\tm 3.9}\\
    \end{tabular}
    \vspace{1ex}
    \caption{\textbf{Classifier retraining.}
    We report all variants of classifier retraining, a subset of which are reported in \Cref{tab:ap_fixed_crt}.
    This strategy improves \oldap, but only mildly affects \fixedap, suggesting that classifier retraining may primarily improve cross-category rankings.
    All results are the average of 3 runs with different random seeds and initializations.
    }
    \label{tab:supp_ap_fixed_pool_crt_all}
\end{table}

\subsection{Evaluating losses for stronger models}
\label{sec:supp_analysis_stronger}

Since certain detector modifications may behave differently as model capacity varies, we re-evaluate the importance of losses and classifier modifications using a stronger model in \Cref{tab:supp_ap_fixed_pool_larger}.
We use a Cascade R-CNN model with a RegNetY-4GF backbone (see \Cref{sec:supp_implementation} for details), which we refer to as the strong model.
We report results using this model in \Cref{tab:supp_ap_fixed_pool_larger}, and compare with the results using a ResNet-50 model (the weak model) reported in \Cref{tab:ap_fixed_pool_all}.
We highlight two key differences.
First, the Federated loss provides clear, significant improvements under both \fixedap and \appool using the strong model (\Cref{tab:supp_ap_fixed_pool_losses}), while it provides only a minor improvement over the weak model under \fixedap (\Cref{tab:ap_fixed_pool_losses}).
This suggests the Federated loss may be more helpful for models with higher capacity.
Second, while the normalized linear layer improves both the weak model and the strong model under all metrics, adding an objectness predictor in addition to the normalized classifier hurts the strong model considerably (dropping from 33.3 to 32.2 under \fixedap).
These results suggest that the normalized classifier is helpful even for high-capacity models, but the objectness predictor is not.

\begin{table*}[t]
    \centering
    \subfloat[\textbf{Loss functions.} With the stronger model, most losses perform roughly equally well under \fixedap and \appool, but the Federated loss shows significant improvements for all metrics.]{%
    \tablestyle{7pt}{1.1}
        \begin{tabular}[t]{r@{\hspace{3ex}}cc@{\hskip 5pt}x{40}}
     Loss & \oldap & \fixedap & \appool \\\midrule
     Softmax CE
             & 28.6 & 31.9 & 32.2 \\\midrule
     Sigmoid BCE
             & 28.5\td{\tm 0.1} & 31.8\td{\tm 0.1} & 32.0\td{\tm 0.2}\\
     EQL \cite{tan2020equalization}
             & 29.4\td{+0.8} & 31.8\td{\tm 0.1} & 32.2\td{+0.0}\\
     Federated \cite{zhou2020cn2}
             & \bf 31.8\td{+3.2} & \bf 33.6\td{+1.7} & \bf 34.7\td{+2.5} \\
     BaGS \cite{li2020overcoming}
             & 30.2\td{+1.6} & 31.9\td{+0.0} & 32.5\td{+0.3}\\
    \end{tabular}
    \label{tab:supp_ap_fixed_pool_losses}
    }
    ~~~~
    \subfloat[\textbf{Classifier modifications.} While the normalized linear layer (`norm') improves both \fixedap and \appool, the objectness predictor (`obj') significantly hurts \fixedap.]{%
    \tablestyle{4pt}{1.1}
    \begin{tabular}[t]{rccc@{\hskip 15pt}cx{50}}
     Loss         &
     Obj & Norm & \oldap & \fixedap & \appool \\\midrule
     \multirow{4}{*}{Softmax CE}
                & \gxmark & \gxmark
                    &     28.6          &     31.9             & 32.2 \\
                & \cmark  & \gxmark
                    &     29.3\td{+0.7} &     31.5\td{\tm 0.4} & 32.4\td{+0.2}\\
                & \gxmark & \cmark
                    &     29.7\td{+1.1} & \bf 33.3\td{+1.4}    & 32.4\td{+0.2}\\
                & \cmark  & \cmark
                    & \bf 30.2\td{+1.6} &     32.2\td{+0.3}    & \bf 32.5\td{+0.3} \\
    \vspace{1.4ex}  %
    \end{tabular}
    \label{tab:supp_ap_fixed_pool_components}
    }
    \vspace{1em}
    \caption{Analyzing losses and components with a higher-capacity detector using a RegNetY-4GF backbone. See \Cref{sec:supp_analysis_stronger} for details.}
    \label{tab:supp_ap_fixed_pool_larger}
\end{table*}

\subsection{Impact of NMS on \fixedap \vs \oldap}
\label{sec:supp_analysis_nms}

Non-max suppression (NMS) is an important, tunable component of modern detection pipelines.
A key parameter in NMS is the intersection-over-union (IoU) threshold: a low-confidence prediction which has IoU greater than this threshold with a high-confidence prediction of the same class is suppressed.
At low thresholds, NMS will ensure predictions are almost entirely non-overlapping by suppressing many predictions.
At high thresholds, by contrast, few predictions are suppressed.
We evaluate the impact of this threshold for NMS for \oldap and \fixedap in~\Cref{tab:supp_nms_comparison}.
Overall, NMS strongly impacts both \oldap and \fixedap significantly.
We highlight one behavior that illustrates a key difference between \oldap and \fixedap.
At high thresholds, where NMS suppresses only a few predictions, \oldap sees a drastic drop in rare class accuracy (\tm 10.1), because predictions for frequent classes crowd out rare class predictions from the common detections per image limit.
However, \fixedap sees a more modest drop in rare class accuracy (\tm 3.3), as \fixedap provides a separate, per-class budget that prevents crowding out of rare class predictions.

\begin{table}
    \tablestyle{2pt}{1.1}
    \centering
    \begin{tabular}{c@{\hskip 4pt}c@{\hskip 5.5ex}c@{\hskip 5.5ex}cc@{\hskip 8pt}c@{\hskip 5.5ex}c@{\hskip 5.5ex}cc}
        \toprule
        NMS & \multicolumn{4}{c}{\oldap} & \multicolumn{4}{c}{\fixedap} \\
        \cmidrule(lr){2-5}  \cmidrule(lr){6-9}
        Thresh & AP & \apr & \apc & \apf & AP & \apr & \apc & \apf \\
        \midrule
        0.5 & 22.7             & 13.3 & 21.2 & 28.6 & 25.5 & 18.8 & 24.8 & 29.1 \\
        \midrule
        0.1 & 23.2\td{+0.5}    & 14.0\td{+0.7} & 22.1 & 28.6
            & 25.1\td{\tm 0.4} & 18.1\td{\tm 0.7} & 24.4 & 28.9 \\
        0.3 & 23.5\td{+0.8}    & 13.9\td{+0.6} & 22.2 & 29.1
            & 25.8\td{+0.3}    & 18.7\td{\tm 0.1} & 25.2 & 29.6 \\
        0.7 & 20.0\td{\tm 2.7} & ~~8.8\td{\tm 4.5} & 18.8 & 26.2
            & 23.7\td{\tm 1.8} & 18.2\td{\tm 0.6} & 23.1 & 26.8 \\
        0.9 & 12.6\td{\tm 10.1}& ~~5.2\td{\tm8.1} & 12.2 & 16.3
            & 16.5\td{\tm 9.0} & 15.5\td{\tm 3.3} & 16.6 & 16.9 \\
        \bottomrule
    \end{tabular}
    \caption{\textbf{Varying NMS thresholds, comparing \oldap and \fixedap.}
    Varying the intersection-over-union threshold for NMS impacts both \oldap and \fixedap, compared to the default threshold of 0.5.
    At the high threshold of 0.9, NMS suppresses very few boxes.
    For \oldap, this results in a significant drop in accuracy for rare classes (\tm 8.1), as overlapping predictions for frequent classes fill up the detections-per-image budget.
    For \fixedap, however, the effect is much smaller (\tm 3.3), due to the independent budget per classes.
    }
    \label{tab:supp_nms_comparison}
\end{table}

\section{Implementation details}
\label{sec:supp_implementation}

\smallsec{Experimental Setup for Tables \ref{tab:artificially_limit_dets_per_class}, \ref{tab:higher_dets_per_im}, \ref{tab:dets_per_cat}.}
For experiments in Tables \ref{tab:artificially_limit_dets_per_class} and \ref{tab:dets_per_cat}, and the LVIS results in Table \ref{tab:higher_dets_per_im},
we closely follow the setup in \Cref{sec:experimental_setup}, but all results are from a single random seed and init.
We follow the same setup for COCO results in Table \ref{tab:higher_dets_per_im} with two modifications: (1) We train for 270k iterations with a 0.1$\times$ learning rate decay at 210k and 250k iterations, and (2) we use a uniform sampler.

\smallsec{Classifier heads: Objectness.}
The objectness predictor in \Cref{tab:ap_fixed_components} is implemented as an additional linear layer parallel with the classifier.
This predictor is trained with sigmoid BCE loss, with a target of $1$ for proposals matched to any object, and $0$ otherwise.
Let $s_\text{obj}$ be the output of the objectness predictor after sigmoid, and $s_c$ be the score for class $c$ from the classifier after softmax.
At test time, we update scores for each class as $s'_c = s_\text{obj} \cdot s_c$.
BaGS uses this same objectness predictor by default, so we do not add a separate objectness predictor.

\smallsec{Classifier heads: Normalized layer.}
We replace the standard classifier with the following, as in~\cite{wang2020seesaw}:
\begin{align*}
  f(x; w, b, \tau) = \frac{\tau}{\norm{w}_2 \norm{x}_2} w^T x + b,
\end{align*}
where $\tau$ is a temperature parameter tuned separately for each loss.
For softmax CE and BaGS, $\tau=20.0$; for sigmoid BCE, Federated, and EQL losses, $\tau=50.0$.
When using an objectness predictor with the normalized classifier, we replace the objectness predictor with a normalized layer as well. \Cref{fig:norm_fc_code} shows a concise PyTorch implementation.

\begin{figure}[t]
\begin{lstlisting}[language=Python]
from torch.nn import functional as F

def forward_normalized(linear, x, temperature):
    """
    Args:
        x (torch.Tensor): Input features, shape (n, d).
        temperature (float): Temperature hyperparam.
        linear (nn.Linear): Standard linear layer.
    """
    x_normed = F.normalize(x, p=2, dim=1)
    w_normed = F.normalize(linear.weight, p=2, dim=1)
    return F.linear(
        temperature * x_normed, w_normed, linear.bias
    )
\end{lstlisting}
\vspace{-1em}
\caption{Pytorch code for implementing the forward pass of a normalized linear layer using a standard linear layer.}
\label{fig:norm_fc_code}
\end{figure}

\smallsec{Classifier retraining.}
For classifer retraining experiments in \Cref{tab:ap_fixed_crt}, we train models in two phases.
In Phase 1, we train the baseline model end-to-end following the setup in \Cref{sec:experimental_setup}, using the Phase 1 sampler.
In Phase 2, we randomly re-initialize the classifier weights and biases of the model. We fine-tune only the classifer weights and biases with the specified Phase 2 sampler
for 90k iters using a minibatch size of 16 images with a 0.1$\times$ learning rate decay applied after 60k and 80k iterations.
The learning rate starts at 0.02 and a weight decay of 1e\tm4 is used.

\smallsec{RegNetY-4GF model.}
For the `RegNetY-4GF' model in \Cref{tab:ap_fixed_pool_larger_models} and \Cref{tab:supp_ap_fixed_pool_larger}, we use a Cascade R-CNN model~\cite{cai2018cascade} with a RegNetY-4GF~\cite{radosavovic2020designing} backbone using FPN~\cite{lin2017feature}.
This model is trained following the \Cref{sec:experimental_setup} setup, with important modifications to achieve high accuracy using the stronger capacity backbone.
The model is trained for 270k iterations, with a 0.1$\times$ learning rate decay applied after 210k and 250k iterations.
The weight decay is set to 5e\tm5 to match the weight decay used for ImageNet pre-training (a standard weight decay of 1e\tm4 decreases \oldap by more than 3 points).
Stronger data augmentation was needed to prevent overfitting for rare categories; we resize the larger size of training images randomly between 400px and 1000px, instead of the default strategy (used for all other experiments) of picking a random scale from 640px to 800px with a step of 32.

\begin{table}[t]
\resizebox{\linewidth}{!}{
\tablestyle{2pt}{1.1}
\begin{tabular}{lc@{\hskip 5pt} lccc}
Loss & Param & Description & Default & Search & Final \\\midrule
EQL & $\lambda$ & Frequency threshold & 1.76e\tm3 & \makecell{\{1e-4, 5e-4, 1e-3,\\ 1.76e-3, 5e-3\}} & 1e\tm3 \\\addlinespace[1ex]
BaGS & $\beta$ & BG sample ratio & 8.0 & \makecell{\{4.0, 8.0, 16.0,\\32.0, 64.0\}} & 16.0 \\\addlinespace[1ex]
Federated & $|S|$ & Neg. classes sampled & 50 & \{10, 50, 100\} & 50 \\
\end{tabular}
}
\label{tab:supp_loss_tuning}
\caption{Parameters tuned for each loss. See \Cref{sec:supp_implementation} for details.}
\end{table}

\smallsec{Losses.}
As EQL~\cite{tan2020equalization} and BaGS~\cite{li2020overcoming} were developed for LVIS v0.5, and the Federated loss~\cite{zhou2020cn2} was tuned for a CenterNet-based detector~\cite{zhou2019objects}, we tune key parameters for each loss in our setting.
We tune parameters using a Mask R-CNN model with ResNet-50, trained on LVIS v1 closely following the setup in \Cref{sec:experimental_setup}, except with a Uniform sampler instead of RFS, and using a single random initialization and seed instead of three.
We choose hyperparameters which optimize \oldap on the LVIS v1 validation set.
We detail these hyperparameters and their optimal choices in \Cref{tab:supp_loss_tuning}.
In addition to these recently proposed losses, we found it necessary to lengthen the warmup schedule for the `Sigmoid BCE' loss for training stability.
The default for most models is a linear warmup schedule starting with a learning rate of 1e\tm3, ramping up for the first $1,000$ iterations.
For Sigmoid BCE, we found it necessary to start with a lower learning rate of 1e\tm4 and ramp up for $10,000$ iterations.
Our Sigmoid BCE implementation first \emph{sums} the BCE loss over all $K \cdot N$ loss evaluations for the $K$ classes (\eg, 1203 in LVIS v1) and $N$ object proposals in a minibatch and then divides this sum by $N$.

\end{appendices}

{\small
\bibliographystyle{ieee_fullname}
\bibliography{egbib}
}

\end{document}